\def\year{2019}\relax
\documentclass[letterpaper]{article} 
\usepackage{aaai19}  
\usepackage{times}  
\usepackage{helvet}  
\usepackage{courier}  
\usepackage{url}  
\usepackage{graphicx}  
\usepackage{url}   
\usepackage{amsfonts}  
\usepackage{nicefrac}  
\usepackage{amsmath}
\usepackage{lipsum}
\usepackage{booktabs}
\usepackage{comment}
\usepackage{enumitem}
\usepackage{bm}
\usepackage{siunitx}

\begin{document}
\title{Fusion Graph Convolutional Networks}
\author{
  Priyesh Vijayan\textsuperscript{1}\thanks{corresponding author: priyesh@cse.iitm.ac.in}, 
  Yash Chandak\textsuperscript{2}, 
  Mitesh M. Khapra\textsuperscript{1}, 
  Srinivasan Parthasarathy\textsuperscript{3} and 
  Balaraman Ravindran\textsuperscript{1} \\
  \textsuperscript{1}Dept. of CSE and Robert Bosch Centre for Data Science and AI \\
  Indian Institute of Technology Madras, Chennai, India\\
  \textsuperscript{2} Dept. of Computer Science,  University of Massachusetts, Amherst, USA\\ 
  \textsuperscript{3} Dept. of CSE and Dept. of Biomedical Informatics, Ohio State University, Ohio, USA\\
    }
\maketitle

\begin{abstract}
Semi-supervised node classification in attributed graphs, i.e., graphs with node features, involves learning to classify unlabeled nodes given a partially labeled graph. Label predictions are made by jointly modeling the node and its' neighborhood features. State-of-the-art models for node classification on such attributed graphs use differentiable recursive functions that enable aggregation and filtering of neighborhood information from multiple hops. In this work, we analyze the representation capacity of these models to regulate information from multiple hops independently. From our analysis, we conclude that these models despite being powerful, have limited representation capacity to capture multi-hop neighborhood information effectively. Further, we also propose a mathematically motivated, yet simple extension to existing graph convolutional networks (GCNs) which has improved representation capacity. We extensively evaluate the proposed model, F-GCN on eight popular datasets from different domains. F-GCN outperforms the state-of-the-art models for semi-supervised learning on six datasets while being extremely competitive on the other two.
\end{abstract}

\section{Introduction}
In many real-life applications such as social networks, citation networks, protein interaction networks, etc., the entities in an environment are not independent but rather influenced by each other through their interactions. Such relational datasets have been popularly modeled as graphs where the entities make up the node, and the edges represent an interaction. The use of graph-based learning algorithms has increasingly gained traction owing to their ability to model structured data. Categorizing such entities requires extracting relational information from their multi-hop neighborhoods and combining that efficiently with their features. Summarizing information from multiple hops is useful in many applications where there exists semantics in local and group level interactions among entities. Thus, defining and finding the significance of neighborhood information over multiple hops becomes an important aspect of the problem.

Traditionally handcrafted features were widely used to capture relational information. Popular methods used mix of count statistics of label distribution \cite{neville2003collective,lu2003link}, relational properties of nodes like degree, centrality scores \cite{gallagher2010leveraging}, and attribute summaries of immediate neighborhood, etc. Limited by manual engineering, these traditional methods only used raw features built from information associated with immediate (first order or one hop) neighbors. 

The recent surge in deep learning has shown promising results in extracting important semantic features and learning good representations for many machine learning tasks. Deep learning models for such relational node classification tasks can be broadly categorized into models that either learn node representation with structural regularizations \cite{perozzi2014deepwalk,wang2016linked} or those that ignore explicit structural regularization and learn to aggregate neighbors' information \cite{hamilton2017inductive,moore2017DCI,kipf2016GCN}. The former is limited to work only in networks that exhibit high homophily, as they enforce the representation of a node and its neighbor to be similar. The later methods, on the other hand, do no make such assumptions.

Initial works \cite{frasconi1998general} on relational feature extraction with neural nets primarily relied on recursive neural nets to process the graph data. Limited by their ability to deal only with directed ordered acyclic graphs, \cite{scarselli2009graph,gori2005new} introduced Graph Neural Networks (GNNs) which used recursive neural nets to propagate information in any general graph iteratively. However, these GNNs were limited to problems where the entire graph can fit into memory. To extend the work to  sequence generation problems on graphs, \cite{li2015gated} adapted GRUs \cite{cho2014learning} in the propagation step. Recently, \cite{moore2017DCI} proposed an LSTM \cite{hochreiter1997long} based sequence embedding model for node classification but it required randomly ordering first hop neighbors' information, thus essentially discarded the topological structure.

To directly deal with the graph's topological structure, \cite{bruna2013spectral} defined convolutional operations in the spectral domain for graph classification tasks, but required computationally expensive eigen decomposition of the graph Laplacian. To reduce this requirement, \cite{defferrard2016convolutional} approximated the higher order relational feature computation with first order Chebyshev polynomials defined on the graph Laplacian. Graph Convolutional Networks (GCNs) \cite{kipf2016GCN} adapted them to semi-supervised node level classification tasks. GCNs simplified Chebyshev Nets by recursively convolving one-hop neighborhood information with a symmetric graph Laplacian. Recently, \cite{hamilton2017inductive} proposed a generic framework called GraphSAGE with multiple neighborhood aggregator functions. GraphSAGE works with a partial (fixed number) neighborhood of nodes to scale to large graphs. GCN and GraphSAGE are the current state-of-the-art approaches for transductive and inductive node classification tasks in graphs with node features. These end-to-end differentiable methods provide impressive results besides being efficient regarding memory and computational requirements.

Herein, we argue that despite their impressive results these models lack the representation capacity to summarize relevant neighborhood information from multiple hops effectively. We support our argument with our analysis of their representation limitations and also provide a solution to alleviate this issue. Below, we list out primary contributions:
\begin{itemize}
    \item To the best of our knowledge, we are the first to analyze and point out that current state-of-the-art graph convolutional nets lack the representation capacity to regulate different neighborhood information independently. 
    \item We also show that these models capture $K$-hop neighborhood information by a $K^{th}$ order binomial. We take advantage of this to build a binomials basis for the K-hop neighborhood space with outputs from different graph layers corresponding to different hop. With the binomial basis, we define a simple linear fusion layer that can capture any required combination of hops for the end task.
    \item We propose F-GCN, a simple extension to GCNs with the proposed fusion layer. We show that the proposed model outperforms the state-of-the-art models on six datasets while being highly competitive on the other two.
\end{itemize} 
\section{Background}
\subsection{Notations}
Let $G=(V,E)$ denote a graph comprising of vertices, $V$, and edges, $E$ with $|V|=N$ respectively. Let $X \in \mathbb{R}^{N \times F}$ denote the nodes' features and $Y \in \mathbb{B}^{N \times L}$ denote the nodes' labels with $F$ and $L$ referring to the number of  features and labels, respectively. Let $A \in \mathbb{R}^{N \times N}$ denote the adjacency matrix representation of the set of edges, $E$ and let $D \in \mathbb{R}^{N \times N}$ denote the diagonal degree matrix defined as $D_{ii} = \sum_j A_{i,j}$. Let $L = I - D^{-\frac{1}{2}}(A)D^{-\frac{1}{2}}$
denote the normalized graph Laplacian and $\hat{L} = (D+I)^{-\frac{1}{2}}(A+I)(D+I)^{-\frac{1}{2}}$ denote the re-normalized Laplacian \cite{kipf2016GCN}. 

In this paper, a Graph Convolutional Network defined to capture $K$-hop information will have $K$ graph convolutional layers with $d$ dimensional outputs, $h_k$ and an final label layer denoted by $h_L$. $h_L = h_K$ if the last convolution is considered as the label layer otherwise, $h_L = h_{K+1}$ . Let $W_k$ denote the weights associated with the layer, $k$ where $W_1 \in \mathbb{R}^{F\times d}$ is the first hidden layer's weights, $W_L \in \mathbb{R}^{d\times L}$ is the label layer's weights and $W_k \in \mathbb{R}^{d\times d}$ is the intermediate layer's weights. Let $\sigma_k$ define the activation function associated with layer, $k$.

\subsection{Graph Convolutional Networks}
Graph Convolutional Networks (GCNs), introduced in \cite{kipf2016GCN}, is a multi-layer convolutional neural network where the convolutions are defined on a graph structure for the problem of semi-supervised node classification. The conventional two-layer GCNs which captures information up to the $2$\textsuperscript{nd} hop neighborhood of a node can be reformulated to capture information up to any arbitrary hop, $K$ as given below in Eqn: \ref{eqn:1}.
\begin{flalign}
\label{eqn:1}
\begin{split}
h_{0} & = X \\
h_{k} & = \sigma_k(\hat{L} h_{k-1} W_k), \hspace{2mm} \forall\ k \in [1, K-1]. \\
Y & = \sigma_K(\hat{L} h_{K-1} W_L)
\end{split}
\end{flalign}

GCN was used for multi-class classification task with ReLU activation function, $\sigma_k=\text{ReLU} ~\forall k \in [1, K-1]$ and a softmax label layer, $\sigma_K=softmax$. We can rewrite the GCN model in terms of $(K-1)$\textsuperscript{th} hop node and neighbor features as below by factoring $\hat{L}$. 
\begin{flalign}
    \label{eqn:anal-gcn}
    \begin{split}
        h_{k} & = \sigma_k ((\hat{D}^{-\frac{1}{2}}I\hat{D}^{-\frac{1}{2}} + \hat{D}^{-\frac{1}{2}}A\hat{D}^{-\frac{1}{2}}) h_{k-1}W_k) \\ 
            & = \sigma_k (SUM(\hat{D}^{-1}h_{k-1},~   \hat{D}^{-\frac{1}{2}}A\hat{D}^{-\frac{1}{2}}h_{k-1})W_k) 
    \end{split}
\end{flalign}

\subsection{GraphSAGE}
Graph Sample and Aggregator (GraphSAGE) proposed in \cite{hamilton2017inductive} consists of 3 models made up of different differentiable neighborhood aggregator functions. GraphSAGE models were defined for the multi-label semi-supervised inductive learning task, \textit{i.e} generalizing to unseen nodes during training. 
Let the function $Aggregate()$ abstractly denote the different aggregator functions in GraphSAGE, specifically $\text{Aggregate} \in$ \{mean, max pooling, LSTM\} and we will refer to these models as GS-MEAN, GS-MAX and GS-LSTM,  respectively. Similar to GCN, GraphSAGE models also recursively combine neighborhood information at each layer of the Neural Network. GraphSAGE has an additional label layer unlike GCN, i. e.,  $h_L = h_{K+1}$. Hence, $\sigma_k=\text{ReLU}  ~\forall k \in [1, K]$ and $\sigma_L=\text{sigmoid}$. Here, the weights $W_{\hat{k}} \in \mathbb{R}^{2d\times d}$.
\begin{flalign}
    \label{eqn:2}
    \begin{split}
        h_{0} & = X \\
        h_{k} & = \sigma_k({\scriptstyle  CONCAT}(h_{k-1}, Aggregate(h_{k-1}, A) W_{\hat{k}})  \\
                  & \forall k \in [1,K] \\
        Y & = \sigma_{K+1}(h_{K} W_{L})
    \end{split}
\end{flalign}
GraphSAGE models, unlike GCN, are defined to work with partial neighborhood information. For each node, these models randomly sample and use only a subset of neighbors from different hops. This choice to work with partial neighborhood information allows them to scale to large graphs but restricts them from capturing the complete neighborhood information. Rather than viewing it as a choice it can also be seen as restriction imposed by the use of Max Pool and LSTM aggregator functions which require fixed input lengths to compute efficiently. Hence, GraphSAGE constraints the neighborhood subgraph of a node to contain a fixed number of neighbors at each hop. 
\section{Analysis of recursive propagation models} \label{Motivation}
In this section, we first provide a unified formulation of GCN and GraphSAGE as recursively computed graph propagation models. Then, we analyze this unified formulation and show that they lack the representation capacity to regulate information from different hops independently. 

\subsection{Unified Recursive Graph Propagation Kernel}
GCN and GraphSAGE differ from each other in terms of their node features, the neighborhood features, and the combination function. These differences can be abstracted to provide a unified formulation as below.
\begin{flalign}
    \label{eqn:gen}
    \begin{split}
        h_k & = \sigma_k(combine(\Omega_k, \Psi_k)W_k) \\
        \Omega_k & = \alpha h_{k-1} \\
        \Psi_k & = F(A) h_{k-1} \\
    \end{split}
\end{flalign}
Where $\Omega_k$ and $\Psi_k$ denote the $(k-1)$\textsuperscript{th} hop node and it's neighbors' features respectively, $\alpha$ denotes the scaling factor for node features, $F(A)$ denotes the neighbors' weights, and $combine$ denotes the mode of combination of node and it's neighbors' features. For brevity, we have made the neighbors' weighting function to be independent of $h_{k-1}$. 

We can view GCN in terms of the components in Eqn: \ref{eqn:gen} with node features, $\Omega_k$ where $\alpha$ = $\hat{D}^{-1}$, neighbors' features, $\Psi_k$ = $F(A)h_{k-1}$ with $F(A)$ = $\hat{D}^{-\frac{1}{2}}I\hat{D}^{-\frac{1}{2}}$ and combining by summation, $combine= \scriptstyle SUM$.

Similarly, GraphSAGE can be seen to combine nodes' features, $\Omega_k$ with $\alpha$=$I$ and different $F(A)$ based neighbors' features by concatenation, $combine$ = $\scriptstyle CONCAT$. Specifically, $F(A)$ = $D^{-1}A$ for GS-MEAN, $F(A)$ = $\scriptstyle CONCAT$ $(C_{i}A) \forall i$ for GS-MAX where $C_{i}$ is a one hot vector with $1$ in the position of the node with the maximum value for the $i$\textsuperscript{th} feature and for GS-LSTM, $F(A)$ is defined by the LSTM gates which randomly orders neighbors and gives weightage for a neighbor in terms of neighbors seen before. 

The concatenation combination (denoted by square braces below) can also be expressed in terms of a summation of node and neighbors features with different weight matrices, $W_k^\omega \in \mathbb{R}^{d\times d}$ and $W_k^\psi \mathbb{R}^{d\times d}$ respectively by appropriately padding zero matrices, ($\bm{0}$) as shown below. 
\begin{flalign}
\label{eqn:concat}
\begin{split}
h_k & = \sigma_k([\alpha h_{k-1}, F(A)h_{k-1}][W_k^\omega, W_k^\psi]) \\ 
h_k & = \sigma_k([\alpha\cdot h_{k-1}W_k^\omega, \bm{0}] + [\bm{0}, F(A)h_{k-1}W_k^\psi])
\end{split}
\end{flalign}

The $\Omega_k$ and $\Psi_k$ terms for CONCAT and SUMMATION combinations are similar if weights are shared in the CONCAT formulation as shown in Eqn: \ref{eqn:sharedconcat} and Eqn: \ref{eqn:sum},  respectively. Weight sharing refers to $W_k^\omega = W_k^\psi = W_k$.
\begin{eqnarray}
\label{eqn:sharedconcat}
h_k = \sigma_k(([\alpha h_{k-1}, 0] + [0, F(A)h_{k-1}])[W_k, W_k]) \\
\label{eqn:sum}
h_k = \sigma_k(\alpha h_{k-1} + F(A)h_{k-1}W_k)
\end{eqnarray}

For brevity of analysis made henceforth, we only consider the summation model to discuss  the limitations of the recursive propagation kernels without losing any generality on the deductions made. Further, we provide another abstraction to the summation formulation as in Eqn: \ref{eqn:sum} by Eqn:  \ref{eqn:genform}. Henceforth, we refer to Eqn: \ref{eqn:genform} as the generic recursive propagation kernel in the upcoming analysis.
\begin{flalign}
\label{eqn:genform}
\begin{split}
\Phi & = (\alpha + F(A)) \\
h_k & = \sigma_k(\Phi h_{k-1}W_k)
\end{split}
\end{flalign}

\subsubsection{\textbf{Lack of independent regulatory paths to different hops}\\}
Though these propagation models can combine information from multiple hops, their formulation restricts them from independently regulating information from different hops. This is a consequence of recursively computing $K$\textsuperscript{th} hop information in terms of $(K$-$1)$\textsuperscript{th} hop information which results in interdependence among weights associated with the different hop information. We can see this below in the recursively expanded unified graph kernel.
 \begin{flalign}
 \begin{split}
 \label{eqn:unify-exp}
 h_K & = \sigma_K(\Phi \cdot ... \sigma_2(\Phi \cdot(\sigma_1(\Phi \cdot h_0 W_1)W_2)... W_K)  
 \end{split}
 \end{flalign}
 
Let's analyze this with an example of a $3$-hop linear kernel with $K$=$3$ and $\sigma_k$ = $I$ which on expansion yields the following equation: 
 \begin{flalign}
 \begin{split}
 \label{eqn:3hop-weights}
h_3 &  = \alpha^3h_0 \Pi_{k=1}^{k=3} W_k + 3\alpha^2F(A) h_0 \Pi_{k=1}^{k=3} W_k \\ 
     &  ~~ + 3\alpha F(A)^2 h_0 \Pi_{k=1}^{k=3} W_k + F(A)^3 h_0 \Pi_{k=1}^{k=3} W_k
 \end{split}
 \end{flalign}
 
This expansion makes it trivial to note that all the weights influence all the different hop information ($h_0, F(A)h_0, F(A)^2h_0$ and $ F(A)^3h_0$) in the model. For example, if we take the case where only first-hop information (just $F(A)$ term) is required, then there exists no combination of $W_k$s that can provide it under the current model. It should be noted that we cannot obtain the $1$\textsuperscript{st} hop information alone by using a $1$-hop kernel as that would also include $0$-hop information, $F(A)^0h_0 = h_0$.

From the above analysis, we can say that these recursive graph kernels have \textit{limited representation capacity} as they cannot capture information from a particular subset of hops without including information from other hops. The limitation of these networks can be attributed to the specific formulation of recursion used to compute output at every layer. As with every layer $k$ of graph convolutional nets, a new information about the $k$\textsuperscript{th} hop is introduced as $\Phi = I \bm{+ F(A)}$ in $h_k=\sigma_k(\Phi h_{k-1}W_k$). More importantly, the output at $k$\textsuperscript{th} layer passes through a series of computations involving later hops ($j > k$) before reaching the last output layer. And also note that this phenomenon happens for previous layers too. Thus, this leads to a lack of independent computation paths to regulate information from any hop without affecting information from later and earlier hops. 


\textbf{\textit{Inclusion of skip connections:}}
Adding the popular skip connection \cite{he2016deep} to these models as in Eqn: \ref{eqn:skip} improves the multi-hop information regulation capacity.
\begin{eqnarray}
 \label{eqn:skip}
 h_k = \sigma_k(\Phi h_{k-1}W_k) + h_{k-1} , \forall k \in [2, K] \\
 \label{eqn:skip_expand}
  h_{k} = \Sigma_{i=1}^{k} \sigma_k(\Phi \cdot h_{i-1}W_{i})
\end{eqnarray}
On recursively expanding the above equation, it can be seen that adding skip connections to a layer, $k$ results in directly adding information from all the lower hops, $i<k$ as shown in Eqn: \ref{eqn:skip_expand}. Unlike Eqn: \ref{eqn:genform} where the output at each layer, $k$ was only dependent on the previous layer, $h_{k-1}$ accounting to only one computational path; now adding skip connections allows for multiple computational paths. As it can be seen that at the $K$\textsuperscript{th} layer the model has the flexibility to select an output from any or all $h_k \forall k<K$. 

However, it can only discard information beyond a particular hop, $k$ and is still not sufficient to individually regulate the importance of information from individual hop as all hops, $i \leq k$ are inter-dependent. Lets us consider the same example of a $3$-hop model as earlier to capture information from $1$\textsuperscript{st} hop alone ignoring the rest. The best, the $3$-hop model with skip connections can do is to learn to ignore information from $2^{nd}$ and $3^{rd}$ hop by setting $W_2$=$W_3$=$0$ and including $h_1$ along with $h_0$. It can be reasoned as before to see that $W_0$ cannot be set to $0$ as $h_1$ depends on the result of $h_0$ thereby having no means to ignore information from $h_0$. This limits the expressive power to efficiently span the entire space of $K^{th}$ order neighborhood information. To summarize, skip connections at best can obtain information up to a particular hop by ignoring information from subsequent hops. The $\scriptstyle CONCAT$ combination can be perceived as linear skip connection as noted by the authors of GraphSAGE.

\textbf{\textit{Inclusion of different weights:}}
As with the CONCAT model, the summation models can also be modified to have different weights to compute node and neighborhood features. We provide the analysis for such models with non-shared weights in the supplementary material. From that analysis, it can be noticed that models with different weights are powerful enough to obtain any particular hop information ignoring the rest and also can obtain information from a continuous subset of hops \textit{i.e} ($i, i+1, \dots i+j$) ignoring the rest. However, including information from two different hops $i$ and $i+j$ with $j<2$, information from all hops between $i$ and $i+j$ will also be included and can't be ignored. 
\section{Proposed Methodology}
In this section, we propose a simple yet effective extension to GCNs by adding a fusion component that allows them to capture multiple hop information effectively. We motivate and propose this component as a solution that will enable these graph kernels to span the entire space of $K$-hop neighborhood. First, we show that the unified kernel is a binomial combination of node and its' neighborhood information. Thus, at each layer $k$, a $k$\textsuperscript{th} hop kernel is computed by a binomial combination. In light of this, we propose a simple fusion layer that learns to linearly combine information from these binomial bases to span the entire $K$-hop space.

\subsection{\textbf{Binomial basis}\\}
The $K$-hop unified propagation kernel defined in Eqn: \ref{eqn:genform} can be rolled out similar to Eqn: \ref{eqn:unify-exp} and be expressed as a $K$\textsuperscript{th} order binomial in terms of node and it's neighbors' features for the linear activation case as given in Eqn: \ref{eqn:Khopsum}.
\begin{flalign}
\label{eqn:Khopsum}
\begin{split}
h_K &= (\alpha I + F(A))^K h_0 \Pi_{k=1}^K W_k\\ 
\end{split}
\end{flalign}

The higher order binomial term in Eqn: \ref{eqn:Khopsum} when expanded assigns different weights to different $F(A)^k h_0$ terms as seen in Eqn: \ref{eqn:3hop-weights}. These weights correspond to the binomial coefficients of the binomial series, $(\alpha I + F(A))^K$. For example, refer to Eqns: \ref{eqn:2hopsum} and \ref{eqn:3hopsum} corresponding to a $2$-hop and $3$-hop kernel with $\alpha=I$ and $W_K=I$ for simplicity. It can be seen that for the $2$-hop kernel the weights are $[1,2,1]$ and for the $3$-hop kernel it is $[1,3,3,1]$. Thus, these recursive propagation kernels combine different hop information weighed by the binomial coefficients.
\begin{eqnarray}
\label{eqn:2hopsum}
h_2 = h_0 + 2F(A)h_0 + F(A)^2h_0 \\ 
\label{eqn:3hopsum}
h_3 = h_0 + 3F(A)h_0 + 3F(A)^2h_0 + 3F(A)^3h_0
\end{eqnarray}

These weights induce a bias on the importance of each hop which again is a limitation of the kernel design. Any such fixed bias over different hops cannot consistently provide good performance across numerous datasets. In the limit of infinite data, we can expect the $W_k$ parameters to correct these scaling factors induced by these biases. However, as with most graph based semi-supervised learning applications where the amount of available labeled data for training is limited, an undesirable bias can result in a sub-optimal model. 

Existing propagation kernels defined over $K$-hop information, extract relational information by performing convolution operations on different $k$-hop neighbors based on their respective $K$\textsuperscript{th} order binomial. As discussed earlier, biasing the importance of information along with recursive weight dependencies hinder the model from learning relevant information from different hops. These limitations constrain the expressive power of these models from spanning the entire space of $K^{th}$ order neighborhood information. Hence, it is restricted to only a subspace of all possible $K^{th}$ order polynomial defined on the neighborhood of nodes.

\subsection{Linear Fusion Component}

To mitigate these issues with existing models, we propose a minimalistic component for these models, a \textbf{fusion component}. This fusion component consists of parameters to combine the information from the binomial basis defined by the different hop information to effectively scale the entire space of a $K^{th}$ order neighborhood. 

We define the fusion component in Eqn: \ref{eqn:basisfunctions} as a linear weighted combination over $K$-hop neighborhood space spanned by the binomial basis, $h_k$s ($[ h_0, \Phi h_0, \Phi^2h_0, \dots, \Phi^K h_0 ]$) with coefficients, $[\theta_0, \theta_1, \theta_2, \dots, \theta_K]$. The $\theta$ coefficients allow the neural network to explicitly learn the optimal combination of information from different hops. As the $h_k$s are binomials, a parameterized linear combination of these binomials can obtain any combination of the individual hop information. 
\begin{flalign}
\label{eqn:basisfunctions}
\begin{split}
y & = \Sigma_{k=0}^{K} h_k \theta_k
\end{split}
\end{flalign}

\subsection{Fusion Graph Convolutional Network}

We propose Fusion Graph Convolutional Network, F-GCN in equations in \ref{eqn:FGCN}. F-GCN is a minimalistic architecture that adds the fusion component defined in Eqn:  \ref{eqn:basisfunctions} to GCN defined in Eqn: \ref{eqn:1}. It can be seen to combine different $k^{th}$ hop information with the fusion component. The fusion component mentioned in the penultimate line of the equations in \ref{eqn:FGCN} fuses label scores from each propagation step. The fused label scores are then normalized to make label prediction.

\begin{flalign}
\label{eqn:FGCN}
\begin{split}
h_{0} & = \sigma_k (XW_{1}) \\
h_{k} & = \sigma_k (\hat{L}h_{k-1}W_{k}) , \hspace{2mm} \forall\ k \in [1, K]. \\
y & = \Sigma_{k=0}^{K} h_k \theta_k \\
\mathcal{L}  & = softmax(y) ~~ or ~~ sigmoid(y) 
\end{split}
\end{flalign}
The dimensions of $h_k$, $\theta_k$, $Y$ , $W_{0}$, $W_{k}$ are $\mathbb{R}^{N\times d}$, $\mathbb{R}^{d\times L}$, $\mathbb{R}^{N\times L}$,  $\mathbb{R}^{F\times d}$, $\mathbb{R}^{d\times d}$ respectively. F-GCN uses ReLU activations and a softmax label layer accompanied by a multi-class cross entropy for multi-class classification problem or a sigmoid layer followed by a binary cross entropy layer for multi-label classification problem. Since predictions are obtained from every hop, we also subject $h_0$ to a non-linear activation function with weights same as $W_1$ from $h_1$. 

The number of parameters in F-GCN is $O(K*d*d + K*d*L)$, where the first term is for GCN and the second is for the fusion component. GraphSAGE models with no shared weights have $O(K*2(d*d))$ or $O(K*2(2d*d))$ for the summation and concat combination respectively which is more than F-GCN as $L<d$ typically. This simple fusion component with fewer parameters provides additional benefits to F-GCN besides explicitly allowing to capture different hop information. It provides for additional direct gradient flow paths to each of the propagation steps allowing it to learn better discriminative features at the lower hops too which also improves its chances of mitigating vanishing gradient. F-GCN can be seen as a multi-resolution architecture which simultaneously looks at information from different resolutions/hops and models the correlations among them. 

The fusion component is similar in spirit to the Chebyshev filters introduced in \cite{defferrard2016convolutional} for complete graph classification task. The primary difference is that the Chebyshev filters learn coefficients to combine Chebyshev polynomials defined over neighborhood information whereas in F-GCN the coefficients of the filter are used to combine different binomials pertaining to different layers. Moreover, an additional difference is that the Chebyshev polynomial basis is not associated with weights $W_k$ to filter $k$\textsuperscript{th} hop information which can potentially enable the model to learn complex non-linear feature basis. F-GCN also enjoys the benefit of the re-normalization trick of GCN that stabilizes the learning to diminish the effect of vanishing or exploding gradient problems associated with training neural networks.

Note that existing attention mechanisms \cite{vaswani2017attention} are typically defined for a positive combination of information. They have a restricted scoring range based on the activation functions used. In our case, since we needed a mechanism that can scale the amount of addition and subtraction of information from different binomial bases, we opted for the simple linear weighted combination layer. 
\section{Experiments}
We run detailed experiments to compare the performance of GCN Vs. GraphSAGE Vs. FGCN across many datasets. The end task is either semi-supervised multi-class or multi-label node classification. Links to code, dataset, and hyper-parameter details will be made available.

\subsection{Datasets}
Experiments were conducted on eight publicly available datasets from social, citation, product, movie and biological domains. In Table: \ref{dataset-stats}, network statistics of these datasets are given. Dataset details are provided below.

\begin{table}[htb]
    \centering
    \caption{Dataset statistics}
    \label{dataset-stats}
    \resizebox{\columnwidth}{!}
    {
    \begin{tabular}{lllllll}
        \hline
        Dataset     & Network  & Nodes & Edges   & Classes & Multi-label & Features \\ 
        \hline
        CORA       & Citation & 2708  & 5429    & 7       & FALSE       & 1433     \\
        CITE    & Citation & 3312  & 4715    & 6       & FALSE       & 3703     \\
        CORA\_ML & Citation & 11881 & 34648   & 79      & TRUE        & 9568     \\
        HUMAN     & Biology  & 56944 & 1612348 & 121     & TRUE        & 50      \\
        BLOG       & Social   & 69814 & 2810844 & 46      & TRUE        & 5413     \\
        FB    & Social   & 6302  & 73374   & 2       & FALSE       & 2        \\
        AMAZON      & Product  & 16553 & 76981   & 2       & FALSE       & 30       \\
        MOVIE   & Movie    & 7155  & 388404  & 20      & TRUE        & 5297 \\
        \hline
    \end{tabular}
    }
\end{table}

\textbf{Biological network}: We use protein-protein interaction (PPI) network of the Human tissues', as presented in GraphSAGE \cite{hamilton2017inductive}. The dataset contains protein interactions from 24 human tissues. Positional gene sets, motif gene sets, and immunology signatures were used as features/attributes of the state. The ultimate task is to predict the gene's functional ontology. 

\textbf{Movie network}: The movielens-2k dataset available as a part of HetRec 2011 workshop \cite{Cantador:RecSys2011} contains a large number of movies. The dataset is an extension of the MovieLens10M dataset with additional movie tags. The graph constructed based on it considers each movie as a node and a common actor as an edge. The goal of the task is to predict the genre of the movies.

\textbf{Product network}: We follow the set-up similar to \cite{moore2017DCI} and consider Amazon DVD co-purchase network. It is a subset of the co-purchase data,  Amazon\_060  \cite{leskovec2016snap}.  The nodes correspond to DVDs and edges are constructed if two DVDs are co-purchased. The DVD genres are treated as DVD features. The task here is to predict whether DVD sales will cross 7500 or not.

\textbf{Social networks}: We consider the BlogCatalog (BLOG) \cite{wang2010discovering} and Facebook (FB) \cite{pfeiffer2015overcoming,moore2017DCI} datasets. The nodes in the BlogCatalog datasets represents the users of a social blog. Each user's blog tags are treated as the attributes of the nodes and relations based on friendship or fan following are represented as edges. The task is to determine the interests of users. Similarly, in the Facebook dataset, the nodes denote the Facebook users, and its corresponding attributes consists of gender and religious view. The task here is to determine the political view of a user. 

\textbf{Citation Networks}: In citation networks, the research articles are the nodes and citations are the edges. We use Cora, Citeseer, and Cora\_ML datasets from this domain. The node attributes for them are the bag-of-words representation of the article. Predicting the research area of the article is the main objective. Unlike most others which are multi-class datasets, Cora-ML is a multi-label classification dataset \cite{coraML}, 


\begin{table*}[!ht]
\centering
\caption{Transductive Experiments}
\label{table-trans}
\begin{tabular}{l| lllll | l}
\hline
         & NODE   & GCN   & GS-MEAN & GS-MAX & GS-LSTM & F-GCN  \\ \hline
CORA      & 60.222 & \textbf{79.039} & 76.821   & 73.272  & 65.730   & \textbf{79.039} \\
CITE      & 65.861 & \textbf{72.991} & 70.967   & 71.390  & 65.751      & 72.266 \\
CORA\_ML  & 40.311 & 63.848 & 62.800   & 53.476  & OOM      & \textbf{63.993} \\
HUMAN     & 41.459 & 62.057 & 63.753   & 65.068  & 64.231   & \textbf{65.538} \\
BLOG      & 37.876 & 34.073 & 39.433   & \textbf{40.275}  & OOM      & 39.069 \\
FB        & 64.683 & 49.762 & 64.127   & 64.571  & 64.619   & \textbf{64.857 }\\
AMAZON    & 63.710 & 61.777 & 68.266   & 70.302  & 68.024   & \textbf{74.097} \\
MOVIE     & 50.712 & 39.059 & 50.557   & 50.569  & OOM      & \textbf{52.021} \\ 
\hline
\textbf{Penalty} & 10.997 & 6.276 & 2.011   & 2.986 & 5.232   & \textbf{ 0.241} \\
\end{tabular}
\end{table*}

\begin{table*}[!ht]
\centering
\caption{Inductive learning experiment with Human dataset: \footnotemark}
\label{table:ind}
\begin{tabular}{llllllll|l}
NODE   & GCN   & GS-MEAN & GS-MAX & GS-LSTM & Mean* & Max* & LSTM* & F-GCN  \\ \hline
44.644 & 85.708 & 79.634   & 78.054  & 87.111   & 59.800  & 60.000 & 61.200  & \textbf{88.942}
\end{tabular}
\end{table*}

\subsection{Experiment Setup}
We report the experiment results for semi-supervised learning on a test set, populated by randomly sampling $20 \%$ of the dataset. We create the training sets by randomly sampling five sets of $10\%$ nodes from the entire graph. Further, $20\%$ of these training nodes are chosen as the validation set. We do not use the validation set for (re)training. It is ensured that these training samples are mutually exclusive from the held out test data. For all transductive experiments, the reported results are an average of these five different training sets. We report results for inductive learning experiment with the same experimental setup as GraphSAGE on the Human PPI network, where the test nodes and validation nodes have no path to the nodes in the training set. 

The hyperparameters for the models are the number of layers of the network (hops), dimensions of the layers, level of dropouts for all layers and L2 regularization, similar to \cite{kipf2016GCN}. We set the same starting learning rate for all the models across all datasets. We train all the models for a maximum of 2000 epochs using Adam \cite{kingma2014adam} with learning rate set to 1e-2. We use a variant of patience method with learning rate annealing for early stopping of the model. Precisely, we train the model for a minimum of 50 epochs and start with the patience of 30 epochs and drop the learning rate and patience by half when the patience runs out (\textit{i.e.,} when the validation loss does not reduce within the patience window). We stop the training when the model consecutively loses patience for two turns. We added all these components to the baseline codes too, and we even observed an improvement up to $25.91\%$ points for GraphSAGE on their dataset.

For hyper-parameter selection, we search for optimal setting on a two-layer deep feedforward neural network with the node attributes (NODE) alone. We then use the same hyper-parameters across all the other models. We row-normalize the node features and initialize the weights with \cite{glorot2010understanding}. Since the percentage of different labels in training samples can be significantly skewed, we weigh the loss for each label inversely proportional to its total fraction like \cite{moore2017DCI}. We use CONCAT operations for all GraphSAGE models as in the original model and also include skip connections for the rest of the models for all kernel layers. 

We ensured that all models have the same setup in terms of the weighted cross entropy loss, the number of layers, dimensions, stopping criteria and dropouts. We evaluate the models on Micro-F1 scores similar to GraphSAGE \cite{hamilton2017inductive}. The best results for models across multiple hops were reported, which were typically 4 hops for Cora\_ML and HUMAN, 3 hops for Amazon, 2 hops for Cora, Citeseer and FB and 1 hop for Movie and Blog datasets. We report results from different hops rather than a fixed number to be fair to baselines that dropped performance on specific datasets with increased hops.

Our implementation is mini-batch trainable, similar to GraphSAGE. We compare our model, F-GCN, against the node only classifier: NODE, GCN, and all the GraphSAGE variants: GS-MEAN, GS-MAX, and GS-LSTM. GS-LSTM model ran out of memory (OOM) for few datasets as mentioned in the table owing to high parameterization.

\subsection{Experiment results}
Among the baselines, GraphSAGE models with more complex aggregator functions and no shared weights for node and neighborhood features significantly outperform the GCN model with no shared weights and limiting scaling factor, $\alpha$. GCN model performs poorly on datasets where the number of edges is high. This is primarily due to the effect of its node scaling factor in these high degree datasets where the nodes' features are heavily under weighed ($\alpha$= $\hat{D}^{-1}$) relative to its' neighbors' information. The effect of node scaling and biased importance is in agreement with the theoretical justification made in \cite{kipf2016GCN} for the design of re-normalized GCNs over the mean model. However their experimental benefits on highly homophilous datasets, Cora and Citeseer were not achievable on other datasets as shown in Table \ref{table-trans}. This can be observed from the datasets: Blog, Amazon, Movie, and FB as it performs poorly than the classifier which only uses the node attributes, NODE by $\approx$ $3$, $3$, $11$ and $15$ percentage points respectively. 

\textbf{F-GCN Vs. NODE:} F-GCN significantly outperforms the NODE model across the board.

\textbf{F-GCN Vs. GCN:} F-GCN improves over GCN by $\approx 3\%$ in the inductive setup and outperforms GCN on six of the eight datasets in the transductive setup while being comparable to the other two. F-GCN has seemed to have learned to avoid the bias induced by the scaling factor by learning to effectively combine the node features along with additional hop information resulting in improved performances over GCN by up to $15$ percentage points in FB. 
In datasets, where GCN's performance dropped below the NODE model, the F-GCN model has recovered the drop in performance and also further improved the performance by $\approx 2\%$ on BLOG, Movie, and FB. \textit{Moreover, with Amazon, we observe a further $10\%$ point improvement over the node only model.} \textit{With the fusion component, the GCN model which performed poorly among the propagation kernels not only obtained a significant boost in performance but also achieved best overall consistent score with a penalty (average difference from the best) as low as $0.241\%$.}

\textbf{FGN Vs GraphSAGE:} F-GCN outperforms GraphSAGE variants across the board on seven of the eight datasets while slightly underperforming on one. GraphSAGE models have higher flexibility compared to GCNs as they have no shared weights. This explains the significant experimental improvement benefited with concatenation as noted in  \cite{hamilton2017inductive}. F-GCN significantly improves over GraphSAGE models in Human and Amazon dataset. In Amazon dataset, though all of GraphSAGE's variants significantly outperformed GCN and NODE models, F-GCN further improved over GraphSAGE by another $\approx 4\%$. GS-MEAN model, similar to GCN improves over GCN and NODE by $\approx 7\%$ and $4\%$ respectively. This can be attributed to not scaling down the node information as with GCN. Despite that, F-GCN manages to further improve over GS-MEAN by $\approx 6\%$. There is no single winner among GraphSAGE models across datasets. Different variants champion in different datasets among them. \textit{Despite having a single simple aggregation function, F-GCN easily champions over all of them combined except on BLOG.} This suggests that the flexibility to independently regulate information is necessary and irrespective of the complex aggregation functions used, the mild lack in representation capacity holds back GraphSAGE from achieving F-GCN's performance. 

\footnotetext{Our training setup gives better results than what was originally reported for GraphSAGE models (Mean*, LSTM*, Max*)}

Overall, F-GCN improves over the state-of-the-art results on six datasets while being extremely competitive on the other two datasets. Though the proposed fusion component, in theory, is an optimal solution for GCNs with linear activations, they also seem to be experimentally beneficial for GCNs with the piece-wise linear ReLU activations too. Such generalization is not unrealistic in practice, as it is often observed that such generalization of insights from a relaxed linear analysis seems to provide significant clarity and potential improvements on the non-linear front \cite{saxe2013exact,kawaguchi2016deep,hardt2016identity,orabona2017training}.

\begin{figure}[!h]
        \centering
        \includegraphics[scale=0.325]{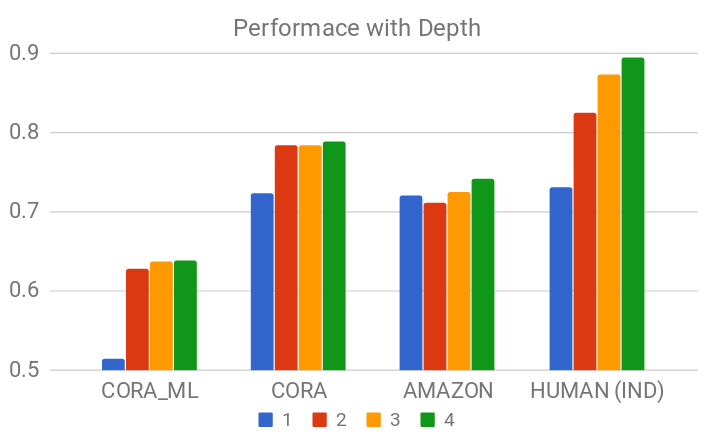}
        \caption{F-GCN performance with hops 1-4}
    \label{fig:depth-perform}
\end{figure}

\textbf{F-GCN robustly captures mutli-hop information }
In real life datasets, there exists a varying amount of information among the interactions between the node and its different distant hop neighbors. An ideal relational model should be able to efficiently capture relevant information while filtering out the increasing noise induced by the expanding neighborhood size with each hop. We demonstrate F-GCN's capability in Fig: \ref{fig:depth-perform} to robustly capture information from multiple hops on different datasets with a varied information pattern. We selected only those datasets that have high relevant relational information for the classification task. These were datasets that obtained significant improvement on results over the node only classifier with just the inclusion of the first hop information.  

In the citation networks (Cora and Cora\_ML), it can be seen that the performances seem to saturate after two hops and with one hop in the Amazon, co-purchase network. Despite that, F-GCN with its capability to selectively regulate information from multiple hops remains unaffected by the noise induced by considering additional hops. 

In contrast, there is a significant increase in performance with the consideration of nodes' higher order neighborhood interactions for the inductive experiment on the protein-protein interaction dataset (HUMAN). In the Human dataset, F-GCN was able to extract relevant information and achieve remarkable performance gain from further hops despite the dataset's high average degree. 

\vspace{-0.45cm}
\section{Conclusion}
We showed that differentiable recursive graph kernels are higher order binomial combinations of node and neighborhood information. Through analysis, we pointed out that such powerful recursively computed binomial functions lack the representation capacity to capture multi-hop neighborhood information effectively. Besides highlighting this critical issue, we also proposed a minimalist fusion component that can alleviate this issue. We empirically demonstrated the effectiveness of coupling the proposed fusion component with GCNs by significantly improving the performance of GCN and achieving highly competitive state-of-the-art results across eight datasets from different domains. For future work, we will incorporate the fusion component with GraphSAGE models and the recent Fast-GCN \cite{chen2018fastgcn}, which samples neighbors to reduce the computation complexity of GCNs.

\bibliographystyle{aaai}
\bibliography{bibliography}
\appendix
\section{Appendix}

\subsection{Inclusion of Different weights}

For convenience, we change the notations for weights associated with the node and the neighbor features to $W_k^0$ and $W_k^1$, respectively. The representation capacity of the recursive graph kernel with different weights for the node and the neighbor features as in Equation \ref{eqn:diffweights} is better than a kernel with shared weights.

\begin{flalign}
\label{eqn:diffweights}
\begin{split}
h_K & = \alpha h_{k-1}W_k^0  + F(A)h_{K-1}W_k^1 \\
\end{split}
\end{flalign}

\begin{figure}[!t]
        \centering
        \includegraphics[scale=0.5]{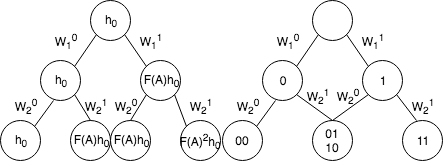}
        \caption{Binomial Computation Trees for Graph Kernels}
    \label{fig:binomtree}
\end{figure}

The flexibility of this formulation can be explained again with the binomial expansion. At every recursive step ($k$), the model can be seen deciding to use only the node information or the neighbors' information or both, by manipulating the weights ($W_k^0$ and $W_k^1$ to be zero or non-zero values). For convenience, we refer the weights to be set if the weights have non-zero values. The decisions of the model can be better understood when visualized as a binomial tree where the nodes are labeled with the computation output, and the edges are labeled by the decision taken, \textit{i.e.} $W_k^0$ or $W_k^1$. The left figure in Figure \ref{fig:binomtree} illustrates the computation graph for a $2$ hop kernel with different weights. 

For any $K$ hop kernel with a $K$ recursive computational layer, there will be $2^K$ unique paths/decisions to make. The $2^K$ paths lead to $2^K$ leaf nodes which compute different $F(A)^k h_0$ terms. $F(A)^k h_0$ terms are available at one or more leaves where the multiplicity of availability is given by the different ways to choose k from K, i.e., $C_k^K$ (binomial coefficient). $h_0$ and $h_K$ terms have only one path ($C_0^K$=$C_K^K$=$1$) whereas the terms $h_k$ ($0<k<K$) have more than one path.

Let string `0' denote the identity transformation, $h_{k-1} W_k^{\bm{0}}$ and `1' denote the F(A) transformation, $h_{k-1} W_k^{\bm{1}}$. We say a transformation has happened if the weights associated with it have non-zero values. To comprehend the dependencies among weights, let us trace the weights along the different paths to leaf nodes. We create the tree on the right in Figure \ref{fig:binomtree} from the output computation tree on the left by relabeling nodes with substrings representing the transformation taken to reach that node. For example, a node labeled `01' indicates that the node was reached by taking an identity transformation followed by an F(A) transformation. Hence, the number of 0s and 1s at each leaf node conveys the pattern to compute each hop information for a $K$ hop kernel. 

In Table \ref{tab:bincounts}, we tabulate the number of identities, $\#W_k^0s$ and the number of $F(A)$ transformations, $\#W_k^1s$ taken to obtain different hop information at the leaves for a $3$-hop kernel. $\#Paths$ in the table denote the number of paths to compute the same. With the example in the table, we generalize the following claims to any $K$ hops.  

\begin{itemize}
    \item $F(A)^k$ computation requires $K-k$ identity transformations ($\#W_k^0 s$) and $k$ $F(A)$ transformations ($\#W_k^1 s$).
    \item All $F(A)^k$ computation has a unique combination of $\#W_k^0 s$ and $\#W_k^1 s$. From this, we can say that the model can learn to obtain any specific hop information without the inclusion of any information from the rest of the hops unlike shared weight models, where all the $F(A)^k$ computations shared the same path.
    \item We cannot independently regulate information from two or more required hops without the inclusion of information from the others hops lying within the range of the required set. This is a consequence of sharing weights among the computation paths as seen in the Figures.
\end{itemize}

Leaving out the first two trivial claims we analyze the last claim.  Let $S$ define the set of all required hops in a $K$ hop graph kernel. Let $i$ and $j$ be the minimum and the maximum hops in that set. For the $i^{th}$ hop, $(K-i)$ $W_k^0 s$ and $i$ $W_k^1 s$ should be set (non zero values) and similarly for the $j^{th}$ hop $(K-j)$ $W_k^0 s$ and $j$ $W_k^1 s$ should be set. $i^{th}$ and $j^{th}$ hop information can be obtained by traversing any path that would satisfy the previously mentioned conditioned on the number of identity and $F(A)$ transformations. Thus, put together $(K-i)$ $W_k^0 s$ and $j$ $W_k^1 s$ will be set to obtain $F(A)^ih_0$ and $F(A)^jh_0$ when traversed along one path in the computation tree, to the leaf from the root. 

Since the model sums up all the leaf nodes, it will also include information from those leaf nodes which can be traversed from the root by following the set $W_k^0$s and $W_k^1$s. We go about the proof by first formulating the condition under which other hops' information can be obtained. Then, we go on to show that if $j > i+1 $ then $j-i$ additional hop information will be included.

Let $y$ denote hops, with $y \notin \{i,j\}$. Computing the $y^{th}$ hop requires $(K-y)$ $W_k^0 s$ and $y$ $W_k^1 s$. $F(A)^y$ can be obtained only if $C_{K-y}^{K-i} \geq 1$ and $C_y^j \geq 1$, which essentially means that $(K-y)$ $W_k^0s$ should be a subset of $K-i$ $W_k^0s$ and $y$ $W_k^1s$ should be a subset of $j$ $W_k^1 s$ which have already been set while considering  $i^{th}$ and $j^{th}$ hop information.

We then find the possible $y$ values under the following three conditions listed below:

\begin{itemize}
    \item $0<y<i$: As $y<i<j$, the required number of $W_k^1 s$ for $y^{th}$ hop is available whereas the required number of $W_k^0 s$ are not as $K-0>K-i-1>K-i$. Hence information from $[0,i)$ is not included.
    \item $j<y<K$: As $y>j$, the required number of $W_k^1 s$ are unavailable though the required number of $W_k^0 s$ can be satisfied as $K-y<K-j<K-i$.
    \item $i<y<j$: As $i<y<j$, the required number of $W_k^1 s$ are satisfied and so is the required number of $W_k^0 s$ as $K-i>K-y>K-j$.
\end{itemize}

This can be clearly seen from the example on the $3^{rd}$ hop kernel presented in Table \ref{tab:bincounts}. When the weights along the unique path for $0^{th}$ and $3^{rd}$ hop information are set, it sets up all the weights. As $0^{th}$ hop information is obtained by doing identity transformation at every layer and $3^{rd}$ hop information is obtained by doing $F(A)$ transformations at every layer, all $W_k^0 s$ and $W_k^1 s$ are set, which would necessarily end up including all the other hop information as all the weights are active. Similarly, it can be shown that when hops $2$ and $3$ are included, no information from hops $0$ and $1$ are included.

\begin{table}[!t]
\centering
\begin{tabular}{l|ll|l}
    & $\#W_k^0 s$ & $\#W_k^1 s$ & $Paths$ \\
\hline
$F(A)^0h_0$ & 3   & 0 & 1 \\ 
$F(A)^1h_0$ & 2   & 1 & 3 \\
$F(A)^2h_0$ & 1   & 2 & 3 \\
$F(A)^3h_0$ & 0   & 3 & 1
\end{tabular}
\caption{Number of Identity and F(A) transformations}
\label{tab:bincounts}
\end{table}


\end{document}